\documentclass{article}

\PassOptionsToPackage{numbers, compress}{natbib}

\usepackage[preprint]{neurips_2019}

\usepackage{neurips_2019}

\usepackage[utf8]{inputenc} 
\usepackage[T1]{fontenc}    
\usepackage{hyperref}       
\usepackage{url}            
\usepackage{booktabs}       
\usepackage{amsfonts}       
\usepackage{nicefrac}       
\usepackage{microtype}      
\usepackage{amsmath, bm, pdfpages, multirow, subfigure,threeparttable}

\title{Adaptive Regularization of Labels}

\author{%
	Qianggang Ding \\
	Tsinghua University\\
	\texttt{dqg18@mails.tsinghua.edu.cn} \\
	\And
	Sifan Wu \\
	Tsinghua University\\
	\texttt{wusf18@mails.tsinghua.edu.cn} \\
	\AND
	Hao Sun \\
	Chinese University of Hong Kong \\
	\texttt{sh018@ie.cuhk.edu.hk} \\
	\And
	Jiadong Guo\thanks{corresponding author.} \\
	PengCheng Laboratory \\
	\texttt{guojd@pcl.ac.cn} \\
	\And
	Shu-Tao Xia \\
	Tsinghua University \\
	\texttt{xiast@sz.tsinghua.edu.cn} \\
}

\begin{document}

\maketitle

\begin{abstract}
Recently, a variety of regularization techniques have been widely applied in deep neural networks, such as dropout, batch normalization, data augmentation, and so on. These methods mainly focus on the regularization of weight parameters to prevent  overfitting effectively. In addition, label regularization techniques such as label smoothing and label disturbance have also been proposed with the motivation of adding a stochastic perturbation to labels. In this paper, we propose a novel adaptive label regularization method, which enables the neural network to learn from the erroneous experience and update the optimal label representation online. On the other hand, compared with knowledge distillation, which learns the correlation of categories using teacher network, our proposed method requires only a minuscule increase in parameters without cumbersome teacher network. Furthermore, we evaluate our method on CIFAR-10/CIFAR-100/ImageNet datasets for image recognition task and AGNews/Yahoo/Yelp-Full datasets for text classification tasks. The empirical results show significant improvement under all experimental settings.
\end{abstract}

\section{Introduction}
\label{sec:introduction}
In recent years, supervised neural networks have been widely used in a variety of deep learning tasks with back-propagation technology. It is well-known that the cross entropy loss function shows remarkable high performances in various tasks in practice. In simple terms, the definition of the cross entropy loss function is the cross entropy between the predicted output of the neural network and the one-hot encoded of labels. The one-hot encoded label is a group of bits among which the legal combinations of values are only those with a single $1$ (represents ground-truth) and all the others $0$. This means that the cross entropy loss based on one-hot encoded labels only focuses on the correctness of the ground-truth category. Malcolm Forbes once said that failure is success if we learn from it. In our approach, we make neural networks learn from previous \textbf{erroneous experience} and benefit future learning. Specifically, our proposed adaptive label regularization enables the neural network to focus not only the correctness but also the incorrectness during the training phase.

Not coincidentally, previous researches \cite{hinton2015distilling, srivastava2015highway, szegedy2016rethinking, zhu2018knowledge, zhang2018deep, lopez2015unifying, pereyra2017regularizing, bagherinezhad2018label} have suggested that the cross entropy loss based on one-hot encoded labels may not be optimal in the classification task. \cite{szegedy2016rethinking, xie2016disturblabel} aimed to regularize the neural network by adding a stochastic perturbation to labels. In addition, \cite{hinton2015distilling} shows that softening one-hot encoded labels could provide more knowledge of the relevance of labels, which refers to as dark knowledge. For example, there exist lots of similarities between the images labeled as "cat" and the images labeled as "dog". The images of both categories contain four legs and one head, while the images labeled as "plane" are significantly different from them. Since one-hot encoded labels are orthogonal each other, they can not indicate the relevance of the labels ("cat" and "dog" are independent of each other). Therefore, the \emph{soft label} was proposed to solve this defect. The essence of the \emph{soft label} is to soften the one-hot encoded labels to vectors of category distribution. But it is very challenging to obtain good soft labels since they are difficult to accurately express using priori knowledge. And another challenge is how to embed soft labels into the training phase of neural networks after getting good soft labels.

To address the above challenges, \cite{hinton2015distilling} indicated that the soft label, \emph{i.e.} the category distribution, was distilled by a cumbersome teacher network using $q_i = \frac{exp(z_i/T)}{\sum_{j=1}^{K} exp(z_j/T)}$, where $K$ is the number of categories and $T$ is a temperature, using a higher value for $T$ produces a softer probability. By learning from soft labels, the student network can compress parameters and even improve performance.

It is worth noting that the existing researches on knowledge distillation \cite{hinton2015distilling, zhu2018knowledge, zhang2018deep, lopez2015unifying} just added the soft loss to the original loss after obtaining a good category distribution. Specifically, they used the KL divergence (equivalent to the cross entropy) between the soft labels and the output of the neural network (soft loss) as part of the loss function in order to supplement the cross entropy with one-hot encoded labels (hard loss). The form is typically as $\mathcal{L} = (1-\alpha) \mathcal{L}_{hard} + \alpha T^2 \mathcal{L}_{soft}$, where $\alpha$ is a trade-off parameter and $T$ is a temperature same as above. But if we think deeply about this combination, we will have some interesting findings. As an example, now given an image classification task which contains three categories such as "cat", "dog", and "plane". Considered a model of this task, the loss function of the model is same as $\mathcal{L}$. Suppose that the soft label of "cat" is $ [0.6, 0.3, 0.1] $ and the neural network model outputs a confidence probability of $ [0.7, 0.2, 0.1] $ for an image labeled as "cat" during the training phase. As we know that the soft loss will push the output of the neural network to the soft label. That is, the soft loss $\mathcal{L}_{soft}$ will \textbf{decrease} the confidence probability of "cat" from $ 0.7 $ to $ 0.6 $. But the hard loss $\mathcal{L}_{soft}$, \emph{i.e.} the cross entropy loss based on the one-hot encoded label, will \textbf{increase} the confidence probability of "cat" from $ 0.7 $ to $ 1 $. Obviously, the optimization goals of the soft loss $\mathcal{L}_{soft}$ and the hard loss $\mathcal{L}_{hard}$ in the loss function $\mathcal{L}$ are \textbf{contradictory}. This unreasonable phenomenon has prompted us to rethink the loss of soft labels, and we will conduct theoretical analysis of it in later sections.

In this work, we propose a novel adaptive label regularization method, which can achieve general improvement in supervised learning tasks. In summary, the contributions of our paper include:

\begin{enumerate}
	\item We define the concept of \emph{residual correlation matrix} and \emph{residual label}, and propose a novel loss function named \emph{residual loss}. Based on these proposed items, our proposed method can not only adaptively regularize the neural network, but also enable the neural network to use erroneous experience in the training phase. 
	\item We show the consistency of residual labels by visualizing the \emph{residual correlation matrix}. And we then analyze the reason why our method can prevent the neural network from overfitting based on the consistency of residual labels.
	\item We perform comprehensive empirical evaluations on five benchmark datasets of two general tasks (image recognition task and text classification task). The empirical results show that our proposed method obtains significant improvement under all settings.
\end{enumerate}

\section{Related work}
 
\textbf{Knowledge Distillation.} The existing researches on knowledge distillation \cite{hinton2015distilling, zhu2018knowledge, zhang2018deep, lopez2015unifying} explored the transmission of the information from deep and cumbersome teacher models to shallow and light student models. The rationale behind this technology is to ensemble the information distilled by one or more teacher models as extra supervision of the student model. \cite{hinton2015distilling} proposed that the soft label distilled by the teacher model implies the information of the label correlation which is known as dark knowledge. \cite{zhang2018deep, zhu2018knowledge} proposed a novel and efficient method of jointly training the teacher model and the student model. Compared to our approach, the methods based on knowledge distillation require training one or more cumbersome teacher models and our method enables the neural network to learn soft labels automatically.

\textbf{Label Regularization.} It is well known that there are various neural network regularization methods like dropout \cite{srivastava2014dropout, zoph2018learning, wan2013regularization, yamada2018shakedrop}, data augmentation \cite{krizhevsky2012imagenet, goodfellow2013maxout, lee2014deeply}, and batch normalization \cite{ioffe2015batch, cooijmans2016recurrent, wu2018group} to improve generalization of neural networks. For label regularization, \cite{szegedy2016rethinking} proposed the label smoothing method redistributes 10 percent of the probability from the ground-truth label to other labels. \cite{xie2016disturblabel} proposed DisturbLabel method sets randomly wrong labels during each training phase. \cite{pereyra2017regularizing} proposed a method to add a confidence penalty regularization term for the outputs of the neural network. It is worth noting that the above label regularization methods do not use the correlation of labels. While our method benefits from the correlation of labels and achieves more significant improvement on the benchmark.

\textbf{Label Refinery.} The label refinery method \cite{bagherinezhad2018label} is an iterative procedure to update ground-truth labels. It shows significant gain using refined labels across a wide range of models. Unlike our method, this method needs to refine the labels several times iteratively. So the time consumption of this approach is much higher than ours.

\section{Proposed method}

In this paper, we consider a $K$-class classifier $ h_\theta(\bm{x}) : \mathcal{X} \to \mathcal{C} $, parameterized by $ \theta $, $ \mathcal{X} $ is the feature space and $\mathcal{C} = \{1,...,K\} $ is the label space. Assuming the classifier $ h = g\circ{f} $ can be decomposed of a embedding function $ f_\theta :\mathcal{X} \to \mathcal{Z} $, which convert input features to logit vectors, and a softmax output layer $ g :\mathcal{Z} \to \mathcal{C} $. The cross entropy loss function is $\mathcal{L}_{CE}(\bm{x},\bm{q}) = - \frac{1}{K} \sum_{i=1}^{K} q_{i} \log p_i$, where $\bm{p} = h_{\theta}(\bm{x}) = softmax(\bm{z})$, $\bm{z} = f_{\theta}(\bm{x})$, and $\bm{q}$ is a one-hot encoded label. 

\subsection{Why not Soft Loss} 
\label{sec:softloss}
Before describing our approach, we first take the knowledge distillation technology as an example to analyze the loss of soft labels (soft loss). In the technology of knowledge distillation, the teacher model distilled a guiding distribution using $q_i^{(soft)}= \frac{exp(z_i/T)}{\sum_{j=1}^{K} exp(z_j/T)}$ and the loss function of knowledge distillation, which we have described in Section \ref{sec:introduction}, typically proposed as $\mathcal{L}_{KD} = (1-\alpha)\mathcal{L}_{hard} + \alpha T^2 \mathcal{L}_{soft}$, where $\mathcal{L}_{hard} = \mathcal{L}_{CE}(\bm{x}, \bm{q})$ and $\mathcal{L}_{soft} = \mathcal{L}_{CE}(\bm{x}, \bm{q}^{(soft)})$. 

By minimizing $\mathcal{L}_{KD}$, the classifier can learn from both the hard label and soft label. It can be considered as a trade-off between $\mathcal{L}_{soft}$ and $\mathcal{L}_{hard}$, in which $\mathcal{L}_{soft}$ contains more dark knowledge than $\mathcal{L}_{hard}$. A balance parameter $\alpha$ can be regarded as the weight between $\mathcal{L}_{hard} $ and $\mathcal{L}_{soft} $. In order to explore the essential process of the optimization, we further compute the gradient of $\mathcal{L}_{KD}$ with respect to $z_i$ (see Appendix A for more details) as follows:
\begin{equation}
\frac{\partial{\mathcal{L}_{KD}}}{\partial{z_i}} = (1-\alpha) \frac{\partial{\mathcal{L}_{hard}}}{\partial{z_i}} +  \alpha T^2 \frac{\partial{\mathcal{L}_{soft}}}{\partial{z_i}} 
\end{equation} 
\begin{equation}
{\rm{where}} \  \frac{\partial{\mathcal{L}_{hard}}}{\partial{z_i}} = p_i - q_i \ \ {\rm{and}} \ \ \frac{\partial{\mathcal{L}_{soft}}}{\partial{z_i}} = \frac{1}{T} (p_i - q_i^{(soft)} ).  
\end{equation}
Since the objective goal of the loss function is to make the gradient $ \frac{\partial{\mathcal{L}_{KD}}}{\partial{z_i}} \to 0$, that is, make $p_i \to \frac{(1-\alpha)q_i + \alpha Tq_i^{(soft)}}{1-\alpha+\alpha T}$, it is easy to learn two drawbacks of $\mathcal{L}_{KD}$: (i) From the above target of $p_i$, we can know that the two parameter $\alpha$ and $T$ both play a trade-off role in $\mathcal{L}_{KD}$. That means the trade-off between the soft label and the hard label is scarcely to be optimal. (ii) Since $\alpha$ and $T$ are fixed during the training phase, it lacks enough flexibility to cope with the situation without given soft labels. That is, the correlation of categories needs to be self-learned under our settings.

\subsection{Adaptive label regularization}
\label{sec:ourmethod}
In practice, the above soft loss obtains excellent performance in various tasks. However, our goal is to explore the regularization of labels in a general neural network without the teacher and take advantage of \textbf{erroneous} knowledge from the previous training experience. Specifically, we propose the \emph{residual loss} in order to regularize output logits of the neural network except for the ground-truth position of them. 

\begin{figure}
	\centering
	\includegraphics[width=12cm]{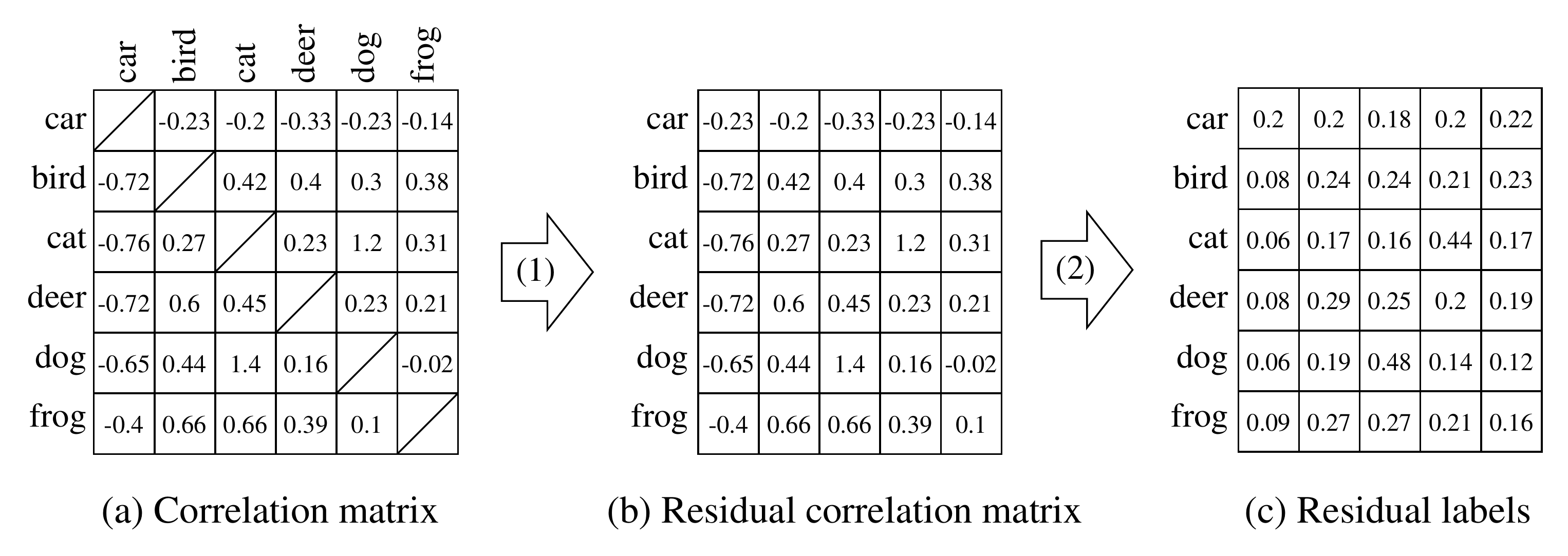} 
	\caption{The relation among correlation matrix, residual correlation matrix, and residual labels: (1) Erasing the $i^{th}$ element of the $i^{th}$ row vector. (2) Softmax normalization. }
	\label{matrix}
\end{figure}

\textbf{Residual Correlation Matrix.} It is known that the correlation matrix $\bm{R} \in \mathbb{R}^{K\times K}$ with the element $r_{i,j}$ indicates the relevance between class $i$ and class $j$. Since we only focus the erroneous knowledge, we define the residual correlation matrix as follows:
\begin{equation}
\bm{S} \in \mathbb{R}^{K\times (K-1)},
\end{equation}
where each row vector $\bm{s}_{i} = (\bm{r}_{i})_{\overline{i}}$ \footnote{We use $(\bm{x})_{\overline{i}}$ to erase the i-$th$ element of $\bm{x}$. }. Figure \ref{matrix} shows the relation between correlation matrix and residual correlation matrix.

\textbf{Residual Label.} \label{reslabel} The residual label $\bm{q}^{(res)}$ is a vector with $K-1$ dimensions, in which each element indicates the probability of samples of category $k$ being wrongly classified into other $K-1$ categories (except for $k$). Since one-hot encoded labels only pay attention to the ground-truth element, residual labels can be a good complement to it. Specifically, we set an embedding layer of which weight is a \emph{residual correlation matrix}. Given a specific label, this embedding layer can convert it to a residual label. Figure \ref{matrix} shows the relation between residual correlation matrix and residual labels.

\textbf{Residual Loss.} To update the residual correlation matrix, assuming at some point during the training phase, the output logits of a min-batch in neural network are $\bm{z}$. We erase the element at the ground-truth position of each row of $\bm{z}$, that is $ z^{(res)}_i = (z_i)_{\overline{k}_i}$ ($i$ represents the $i^{th}$ sample in min-batch). Thus the probability distribution $\bm{p}^{(res)}=softmax(\bm{z}^{(res)})$ represents the erroneous probabilities of all samples in this min-batch. Then we use the cross entropy loss between $\hat{\bm{p}}^{(res)}$ \footnote{We use $\hat{p}$ to block the gradient of $p$ in back-propagation.} and $\bm{q}^{(res)}$ as update loss $\mathcal{L}_{upd}$ to update residual labels. To regularize the output logits, we use the reverse cross entropy loss between $\bm{p}^{(res)}$ and $\hat{\bm{q}}^{(res)}$ as residual loss $\mathcal{L}_{res}$, which aims to transfer the erroneous knowledge to the backbone network. Finally, we can yield a basic combination of these two losses and cross entropy loss of one-hot encoded labels as our total loss:
\begin{equation}\label{equ:lossupd}
\mathcal{L}_{upd} = - \frac{1}{K-1} \sum_{i=1}^{K-1} \hat{p}_i^{(res)} \log q_i^{(res)},
\end{equation}
\begin{equation}
\mathcal{L}_{res} = - \frac{1}{K-1} \sum_{i=1}^{K-1} \hat{q}_i^{(res)} \log p_i^{(res)},
\end{equation}
\begin{equation}
\mathcal{L}_{tot} = \mathcal{L}_{hard} + \mathcal{L}_{res} + \mathcal{L}_{upd}.
\end{equation}

\begin{figure}
	\centering
	\includegraphics[height=8.5cm]{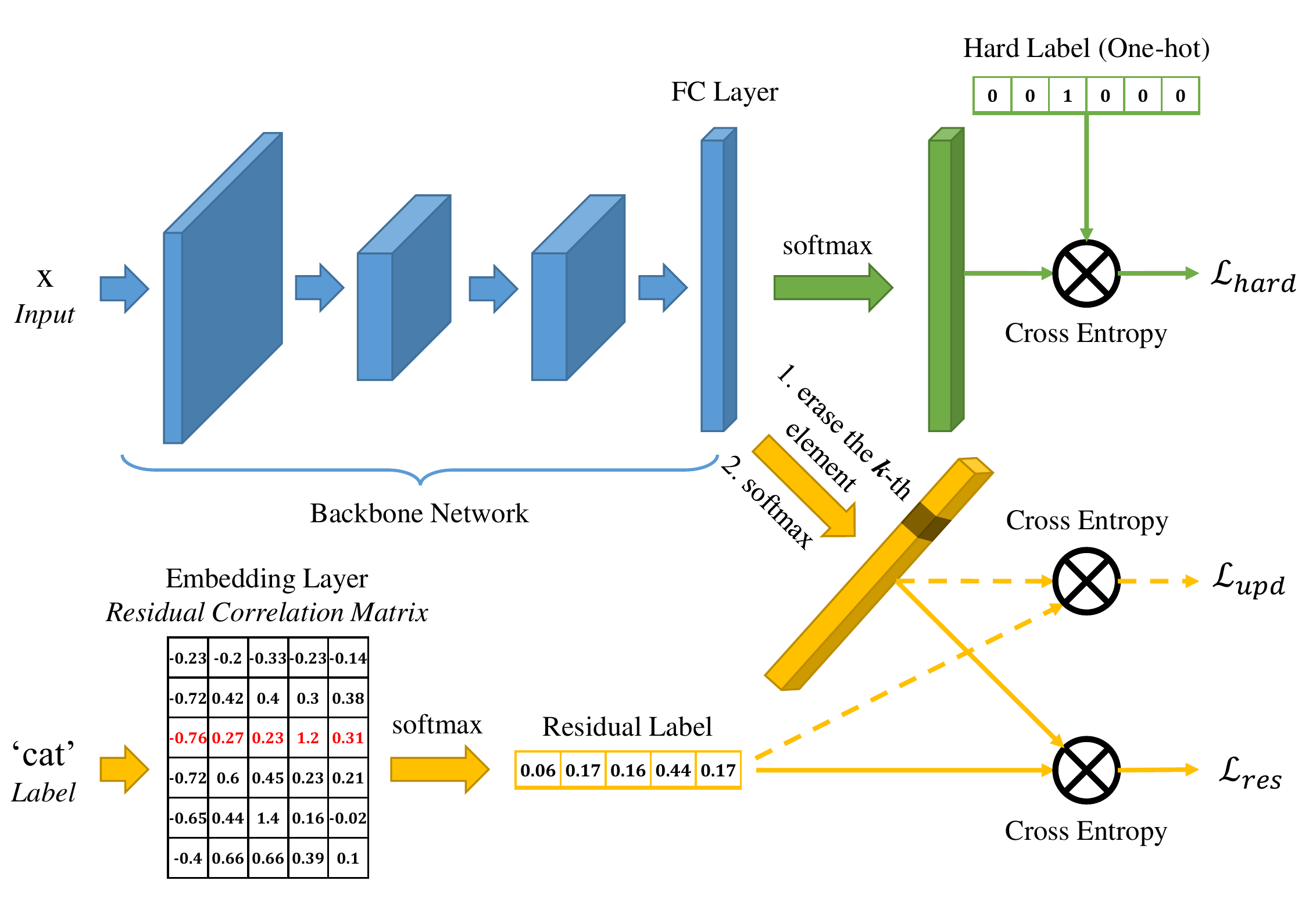} 
	\caption{Our proposed adaptive label regularization pipeline: the part in blue is backbone network and the part in green is the original cross entropy loss. Our proposed part in yellow gives extra regularization of labels to the neural network. Original labels are converted to residual label through an embedding layer, of which weight is a residual correlation matrix. Then we use the cross entropy between the residual label and the output of the neural network to update both the embedding layer and the backbone network.}
	\label{backbone}
\end{figure}

The performance of deep neural networks is well-known to be sensitive to the setting of their 
hyper-parameters. Compared with the soft loss, which contains two hyper-parameters, $\alpha$ and $T$ , our loss function has no hyper-parameter that needed to be manually adjusted. For our backbone network (the part in blue in Figure \ref{backbone}), the objective function is to minimize $\mathcal{L}_{b} = \mathcal{L}_{hard} + \mathcal{L}_{res}$, We further compute the gradient of $\mathcal{L}_{b}$ with respect to $z_i$:
\begin{equation}
\frac{\partial{\mathcal{L}_{t}}}{\partial{z_i}} = 
\left\{
\begin{aligned}
& \frac{\partial{\mathcal{L}_{hard}}}{\partial{z_i}}   &     i=k  ,\\
& \frac{\partial{\mathcal{L}_{hard}}}{\partial{z_i}} + \frac{\partial{\mathcal{L}_{res}}}{\partial{z_i}}   &  i\not = k,
\end{aligned}
\right.
\end{equation}
\begin{equation}
{\rm{where}} \ \  \frac{\partial{\mathcal{L}_{hard}}}{\partial{z_i}} = p_i - q_i  \ \ {\rm{and}} \ \ 
\frac{\partial{\mathcal{L}_{res}}}{\partial{z_i}} = p_i^{(res)} - q_i^{(res)}.
\end{equation}
Intuitively, The optimization target of minimizing $\mathcal{L}_{res}$ is equivalent to decrease the gradient $\frac{\partial{\mathcal{L}_{t}}}{\partial{z_i}}$ to near-zero. So we can obtain from above: (i) When $i=k$, the gradient $\frac{\partial{\mathcal{L}_{t}}}{\partial{z_i}}$ is same as $\frac{\partial{\mathcal{L}_{hard}}}{\partial{z_i}} = p_i - q_i$, decreasing $ p_k - q_k$ equivalent to enforce $p_k$ to be up to 1 because of $q_k=1$. (ii) When $i \not = k$, the gradient $\frac{\partial{\mathcal{L}_{t}}}{\partial{z_i}}$ is the sum of $ \frac{\partial{\mathcal{L}_{hard}}}{\partial{z_i}}$ and $\frac{\partial{\mathcal{L}_{res}}}{\partial{z_i}}$. Since $p_i=0$ with $i\not = k$, the former enforces $p_i$ with $i \not =k$ as small as possible to near-zero and the latter enforces $p_i^{(res)}$ be close to $q_i^{(res)}$. That means, the optimal goal is to let $\bm{p}^{(res)}$ distribution and the residual label $\bm{q}^{(res)}$ as close as possible in the case that the residual is as small as possible. Figure \ref{backbone} shows the pipeline of our proposed method.

Based on the point that the more converged the neural network, the weaker the knowledge gained from the erroneous experience, we set a weight $(1 - \rm{acc}_{train})$ to $\mathcal{L}_{res}$ approximately. Thus the final form of our total loss is $\mathcal{L}_{tot} = \mathcal{L}_{hard} + (1-\rm{acc}_{train})\mathcal{L}_{res} + \mathcal{L}_{upd}$.

\textbf{Initialization.} Since we have no idea about the relevance of categories before training the neural network. Based on the principle of maximum entropy \cite{jaynes1957information}, we assume a uniform distribution to the residual label of each class. That is, each element of the residual correlation matrix would be initialized as the same value, such as zero. Then the softmax output of each row vector of the residual correlation matrix is a uniform distribution. 

Suppose that we make all elements of the residual correlation matrix fixed during the training phase, that means throwing $\mathcal{L}_{upd}$ out from $\mathcal{L}_{tot}$, then the optimization target of $\mathcal{L}_{res}$ is to uniform all residual labels. Intuitively, we can obtain that the optimization target of $\mathcal{L}_{res}$ is same as that of $\mathcal{L}_{hard}$ since the one-hot encoded vector after erasing the ground-truth element is filled with zero, which is also a uniform distribution. Therefore, we can learn that $\mathcal{L}_{upd}$ has a role to regularize the neural network from \textbf{adaptive} residual labels.

\textbf{Combination with label smoothing.} It is worth mentioning that our approach can be well combined with label smoothing regularization. Label smoothing proposed a mechanism for encouraging the model to be less confident about ground-truth to improve generalization. Typically, it gives 10\% confidence to the erroneous classes uniformly, whereas the residual loss $\mathcal{L}_{upd}$ in our approach is to make the distribution of erroneous classes uneven exactly. This means our regularization can complement label smoothing regularization very well, which has been proved by experiments in the next section. 

\section{Empirical results}

In principle, our method can be applied to any tasks based on one-hot encoded vectors. For showing our generalization, we focus on two general tasks (image recognition task and text classification task). We evaluate our method on CIFAR-10, CIFAR-100, and ImageNet-12 for image recognition. For text classification, we evaluate on AGNews, Yahoo! Answer and Yelp Review Full datasets. We conduct all experiments on NVIDIA Tesla V100 (32 GB onboard memory) GPU.

\subsection{Image recognition}

\textbf{Datasets.} We use three benchmark datasets of image recognition in our evaluations: (i) CIFAR-10 \cite{krizhevsky2009learning}: A dataset consisting of $60,000$ images with $32*32$ pixel. It has $10$ classes, of which each class has $50,000$/$10,000$ samples in train/test set. (ii) CIFAR-100  \cite{krizhevsky2009learning}: Similar to CIFAR-10, but it consists of $100$ classes, of which each class has $500$/$100$ samples in train/test set. (iii) ImageNet-12 \cite{russakovsky2015imagenet}: A huge image recognition dataset from ILSVRC 2012, which consisting of more than $14$ million samples in $1,000$ classes. 

\textbf{Settings.} For CIFAR-10 and CIFAR-100 datasets, we use ResNet-18 \cite{he2016identity} with pre-activation and WideResNet-28-10 \cite{zagoruyko2016wide} as backbone network architectures. We use SGD optimizer with Nesterov momentum \cite{sutskever2013importance} and set initial learning rate to $0.1$, momentum to $0.9$ and mini-batch size to $128$. The learning rate dropped by $0.1$ at the $60/120/160^{th}$ epochs and we train for $300$ epochs. We also set data augmentation such as horizontal flips and random crops to samples during the training stage. For ImageNet-12 dataset, we use a modified version of ResNet-50-v1, ResNet-101-v1,and ResNet-152-v1 \cite{he2016deep} as backbone networks to evaluate our method. The difference between the original ResNet-50-v1 network \cite{he2016deep} and ours is that a stride $2$ is used on the Conv-3x3 layer rather than the first Conv-1x1 in the bottleneck. In addition, we use mixed precision training \cite{micikevicius2017mixed} in ImageNet experiment which offers significant computational speedup.

\textbf{Results on CIFAR.} The results on two CIFAR datasets are shown in Table \ref{cifar-table}. We compare our proposed adaptive label regularization (ALR) method, label smoothing regularization (SLR) method and the combination of ALR and SLR methods (ALR-S), of which the best results are marked in bold (the same as below). As we can see, our proposed ALR and ALR-S methods improve the performance of the baselines under all settings. In addition, it is worth mentioning that ALR can complement LSR very well (ALR-S), which achieves the accuracy enlargement of $1.54\%$ from the baseline under the setting of ResNet-18 model on CIFAR-100 dataset.

\textbf{Results on ImageNet.} The results on ImageNet-12 dataset are shown in Table \ref{imagenet-table}. We still compare SLR method, ALR method, and ALR-S. As we can see, our proposed ALR and ALR-S methods both can improve all the top-1 accuracy of the baselines, but the top-5 accuracy of our methods is slightly lower than it of the baselines. In our point, our regularization of the probability of erroneous makes the correlation of the top-5 predicted categories high. That is, if the neural network mispredicts a cat into an orange, the top-5 predicted category will be similar categories to "orange" like "lemon" and "apple". So the prediction will be more random, which leads to a smaller probability to predict as "cat" than the baselines.

\begin{table}
	\small
	\caption{Results on CIFAR-10/CIFAR-100. (\%) (results averaged over 5 runs)}
	\label{cifar-table}
	\centering
	\begin{tabular}{llllll}
		\toprule
		\multicolumn{2}{l}{Datasets} & \multicolumn{2}{c}{CIFAR-10} & \multicolumn{2}{c}{CIFAR-100}            \\
		\midrule
		Model                         & Method         & Accuracy     & Params  & Accuracy & Params \\
		\midrule
		\multirow{4}{*}{ResNet18}     & Baseline       & $95.28\pm0.21$    & 11.17M &  $77.54\pm0.31$  &  11.22M \\
		& SLR            &   $95.41\pm0.10$        & 11.17M  &    $78.24\pm0.26$     & 11.22M \\
		& ALR         &   $95.42\pm0.16$    & 11.17M + 0.1K  &   $78.15\pm0.27$   & 11.22M + 10K \\
		& ALR-S          &   $\bm{95.56\pm0.13}$    & 11.17M + 0.1K   &   $\bm{79.08\pm0.24}$  & 11.22M + 10K \\
		\midrule
		\multirow{4}{*}{WideResNet-28-10}    & Baseline   &   $96.13\pm0.08$    &  36.48M  & $81.20\pm0.08$   & 36.54M  \\
		& SLR            &  $96.18\pm0.12$    & 36.48M      &  $81.13\pm0.11$     & 36.54M     \\
		& ALR           &  $96.35\pm0.11$     & 36.48M + 0.1K  &  $\bm{81.31\pm0.14}$    & 36.54M + 10K  \\
		& ALR-S      &  $\bm{96.37\pm0.13}$   & 36.48M + 0.1K  & $81.28\pm0.12$      &  36.54M + 10K  \\
		\bottomrule
	\end{tabular}
\end{table}

\begin{table}
	\small
	\caption{Results on ImageNet-12. (\%)}
	\label{imagenet-table}
	\centering
	\begin{tabular}{lllllll}
		\toprule
		\multirow{2}{*}{Method}  & \multicolumn{2}{c}{ResNet-50} & \multicolumn{2}{c}{ResNet-101} & \multicolumn{2}{c}{ResNet-152} \\
		\cmidrule(){2-7}
		         & Top-1 Acc.     & Top-5 Acc.     & Top-1 Acc.      & Top-5 Acc.     & Top-1 Acc.      & Top-5 Acc.     \\
		\midrule
		Baseline     &  $76.17$        &  $93.03$        &  $77.53$         & $93.82$        & $77.71$          & $93.89$              \\
		SLR          &  $76.25$        &  $\bm{93.24}$        &  $77.66$         &  $\bm{93.93}$        & $77.93$          & $94.02$              \\
		ALR          &  $76.59$        &  $92.73$        &  $77.82$         &  $93.63$        & $78.54$         &  $93.80$        \\
		ALR-S        &  $\bm{76.74}$        &  $92.89$        &  $\bm{77.96}$         &  $93.48$        & $\bm{78.68}$          & $\bm{94.06}$   \\
		\bottomrule
	\end{tabular}
\end{table}

\subsection{Text classification}

\textbf{Datasets.} We use three benchmark datasets of text classification in our evaluations: (i) AGNews \cite{del2005ranking}: A dataset of topic classification over four categories of Internet news articles such as World, Entertainment, Sports, and Business. It consists of $30,000/1,900$ records in train/test set. (ii) Yahoo! Answers: A topic classification dataset with ten categories obtained from Yahoo! Webscope program, of which each category has $140,000/5,000$ records in train/test set. (iii) Yelp Review Full: A dataset obtained from Yelp Dataset Challenge in 2015, the task of which is sentiment classification of polarity star labels ranging from $1$ to $5$. There have $130,000/10,000$ samples in train/test set in each star.

\textbf{Settings.} For all datasets, we evaluate four methods of text classification with or without our proposed regularization, including FastText \cite{joulin2017bag}, TextRNN \cite{lai2015recurrent}, CharCNN \cite{zhang2015character-level}, and Transformer \cite{vaswani2017attention}. We follow the experimental setting of the original papers of these methods. We use Adam optimizer and set initial learning rate to $0.0001$, mini-batch size to $128$. The learning rate dropped by $0.5$ every $10$ epochs and we train for $30$ epochs.

\textbf{Results.} The results are shown in Table \ref{text-table}. As we can see, our proposed adaptive label regularization can be generalized to improve the performance under all settings no matter which method the network architecture is based on. These empirical results make us believe that our method can widely improve the performance of neural networks in any tasks based on one-hot encoded labels.

\begin{table}
	\small
	\caption{Results of text classification. Metric: Acc. (\%)}
	\label{text-table}
	\centering
	\begin{tabular}{lllllll}
		\toprule
		Datasets & \multicolumn{2}{c}{AGNews} & \multicolumn{2}{c}{Yahoo! Answers} & \multicolumn{2}{c}{Yelp Review Full} \\
		\midrule
		Method & Baseline       & ALR-S       & Baseline              & ALR-S              & Baseline       & ALR-S       \\
		\midrule
		FastText             &  $88.97$   &   $\bm{89.28}$  & $70.58$ & $\bm{71.25}$  & $59.34$  &  $\bm{59.98}$  \\
		TextRNN              &  $88.85$   &   $\bm{89.57}$  & $68.87$ & $\bm{70.33}$  & $57.19$  &  $\bm{58.12}$  \\
		CharCNN              &  $85.78$   &   $\bm{87.91}$  & $70.24$ & $\bm{71.50}$  & $60.72$  &  $\bm{60.98}$  \\
		Transformer          &  $89.51$   &   $\bm{90.29}$  & $70.97$ & $\bm{71.24}$  & $61.50$  &  $\bm{62.32}$  \\
		\bottomrule
	\end{tabular}
\end{table}

\subsection{Comparison with DML}

In addition, we compare our method with Deep Mutual Learning (DML) \cite{zhang2018deep}, which is a representative online method of knowledge distillation. For Deep Mutual Learning method, we use a combination of two sub-network with the same architecture (two ResNet-32 sub-networks or two WidResNet-28-10 sub-networks). For the general knowledge distillation method, we transfer the knowledge from WideResNet-28-10 network to ResNet-32 network. In this experiment, we evaluate all the methods on CIFAR-100 dataset. For fairly comparison, we follow the experimental setting of \cite{zhang2018deep} , which set the mini-batch size to $64$, and drop the learning rate by $0.1$ every $60$ epochs. The results are shown in Table \ref{dml-table}. It is worth noting that our method can achieve nearly the same performance as DML, but only uses half of the parameters. 

\begin{table}
	\small
	\caption{Results of comparison with DML on CIFAR-100. Metric: Acc. (\%). "*": reported results.}
	\label{dml-table}
	\centering
	\begin{tabular}{lcccc}
		\toprule
		Network  & \multicolumn{2}{c}{ResNet-32} & \multicolumn{2}{c}{WideResNet-28-10} \\
		\midrule
		Method   & Accuracy       & Params       & Accuracy              & Params              \\
		\midrule
		Baseline   &   $68.99^*$    &  $1\times$     & $78.69^*$     & $1\times$  \\
		Knowledge Distillation     &  $69.48^*$   &  $\sim3\times$  &  $-$    & $-$  \\
		DML       &  $70.93^*$     &   $\sim2\times$  &  $80.18^*$     & $\sim2\times$ \\
		ALR-S      &    $\bm{70.97\pm0.37}$      &  $\sim1\times$ &  $\bm{80.70\pm0.18}$    &   $\sim1\times$  \\
		\bottomrule
	\end{tabular}
\end{table}

\begin{figure}[htbp]
	\centering
	\subfigure[1 epoch ($\mu = 44.12\%$)]{
		\includegraphics[width=1.7in]{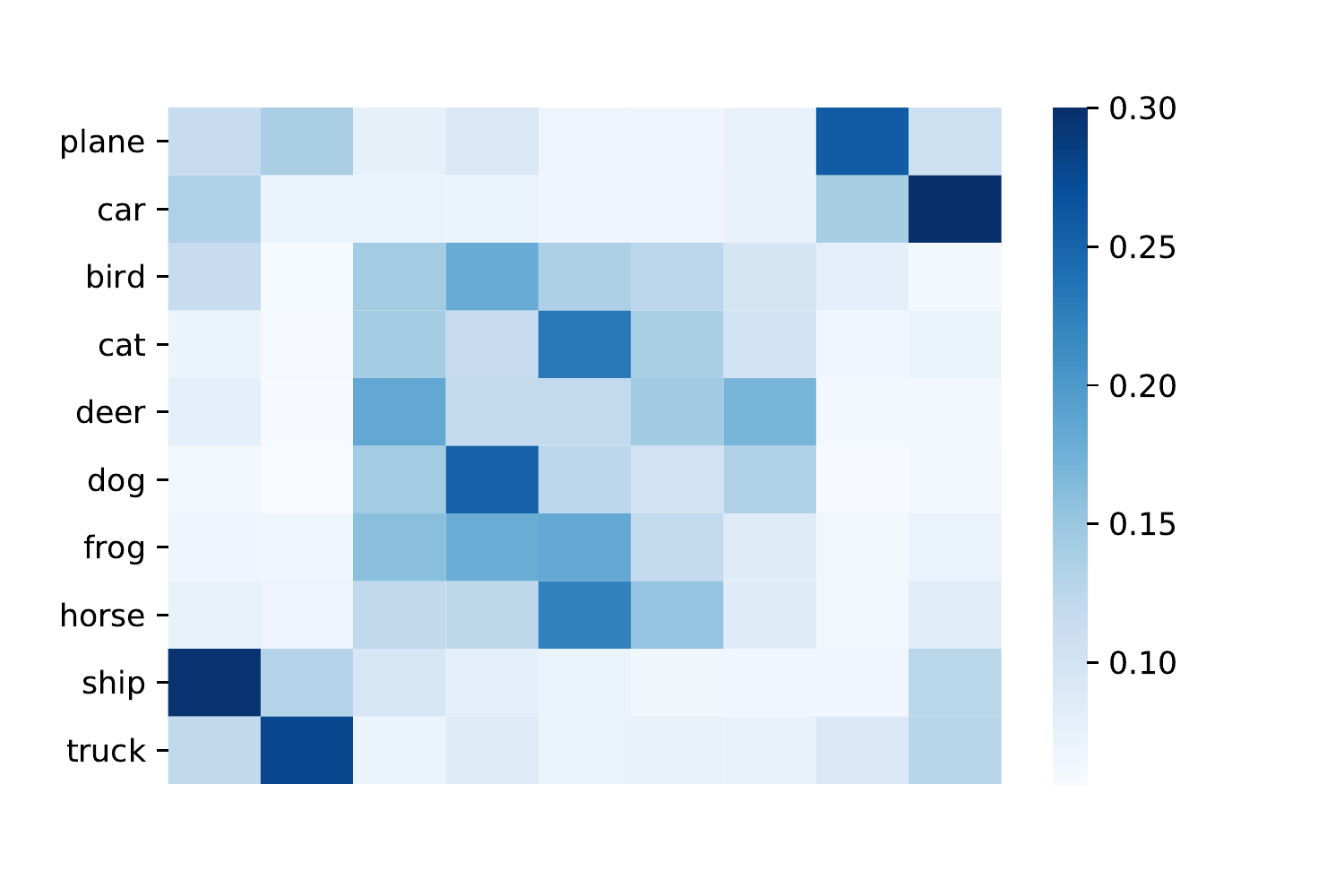}
	}
	\subfigure[5 epochs ($\mu = 80.21\%$)]{
		\includegraphics[width=1.7in]{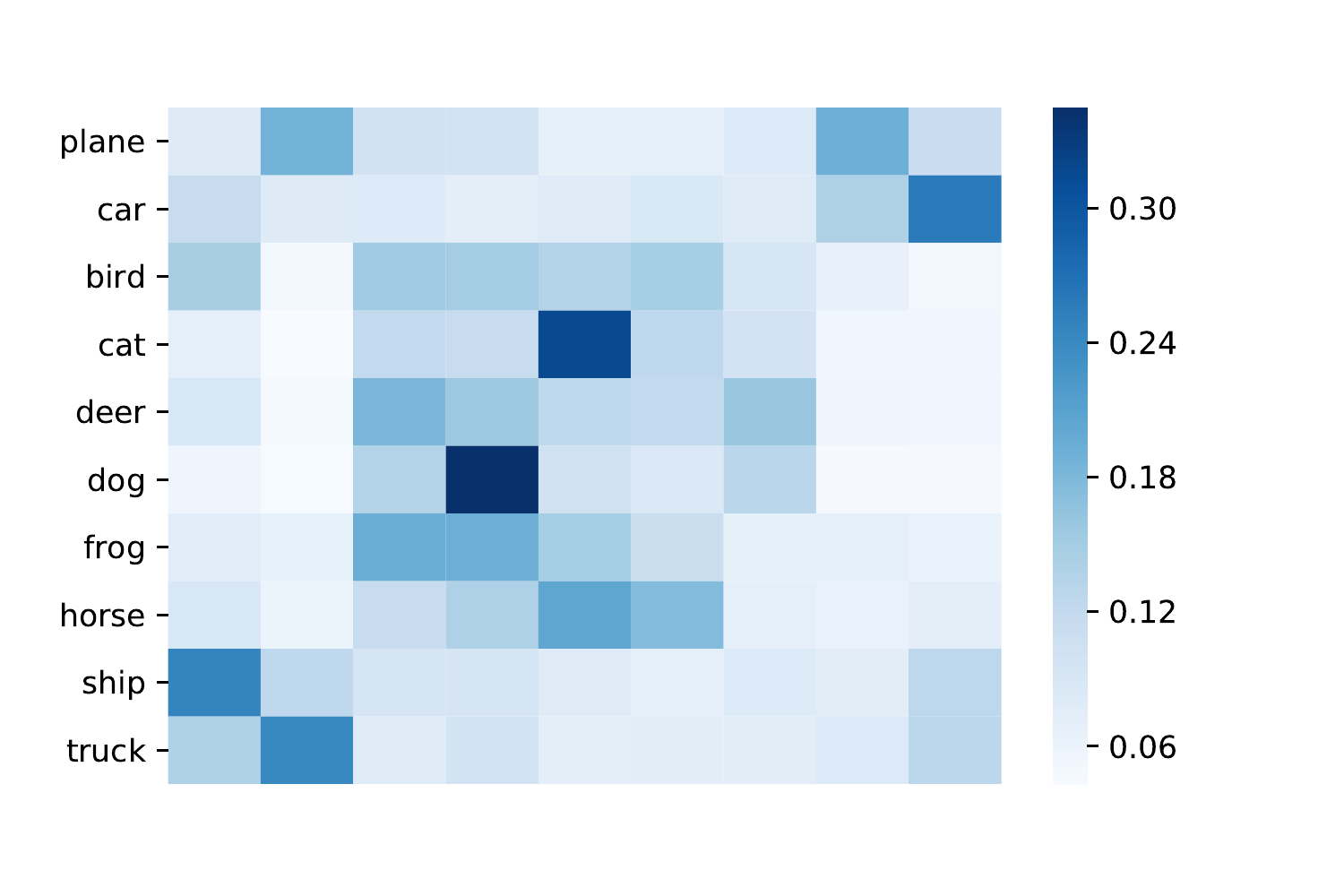}
	}
	\subfigure[10 epochs ($\mu = 84.60\%$)]{
		\includegraphics[width=1.7in]{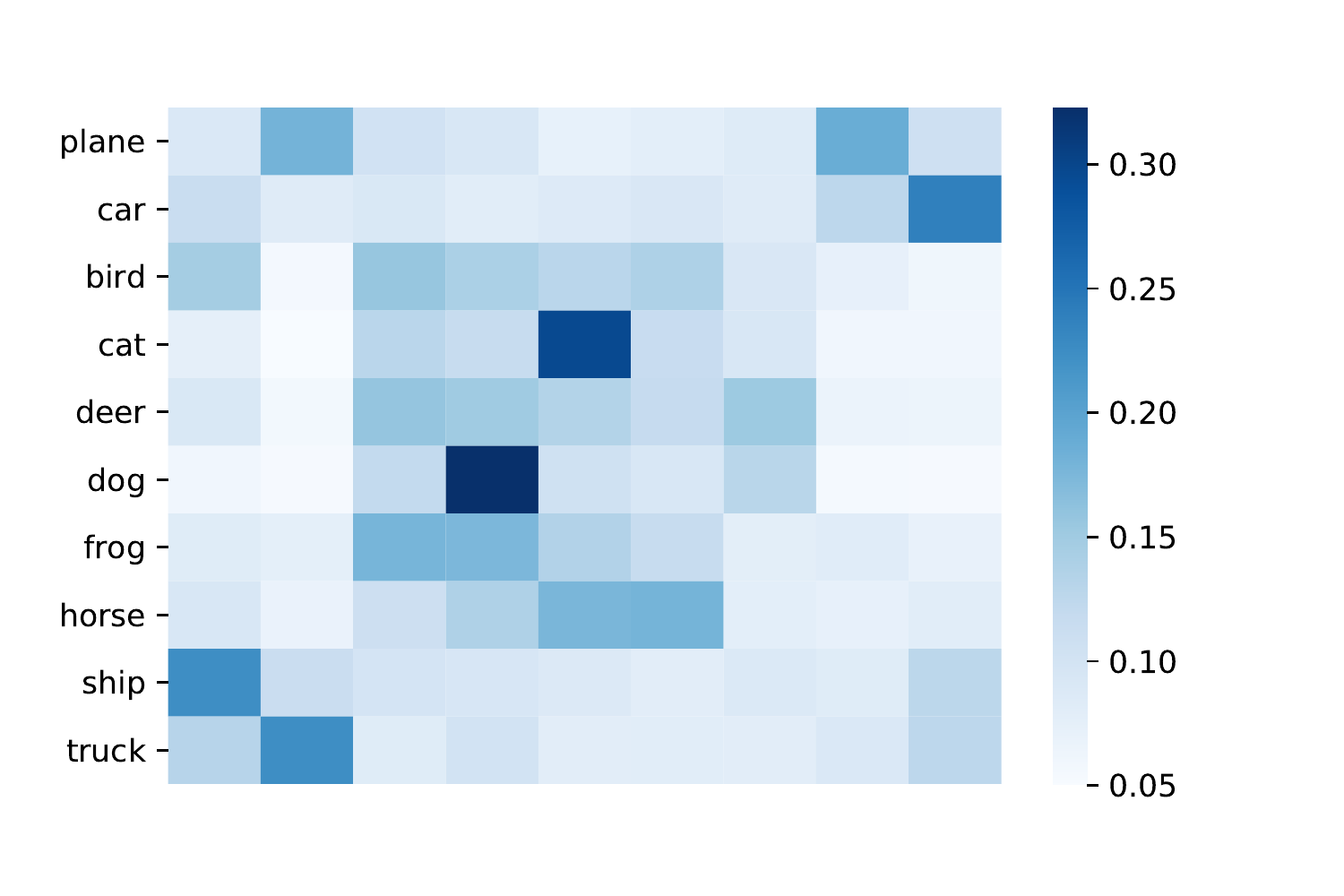}
	}
	\quad    
	\subfigure[100 epochs ($\mu = 96.94\%$)]{
		\includegraphics[width=1.7in]{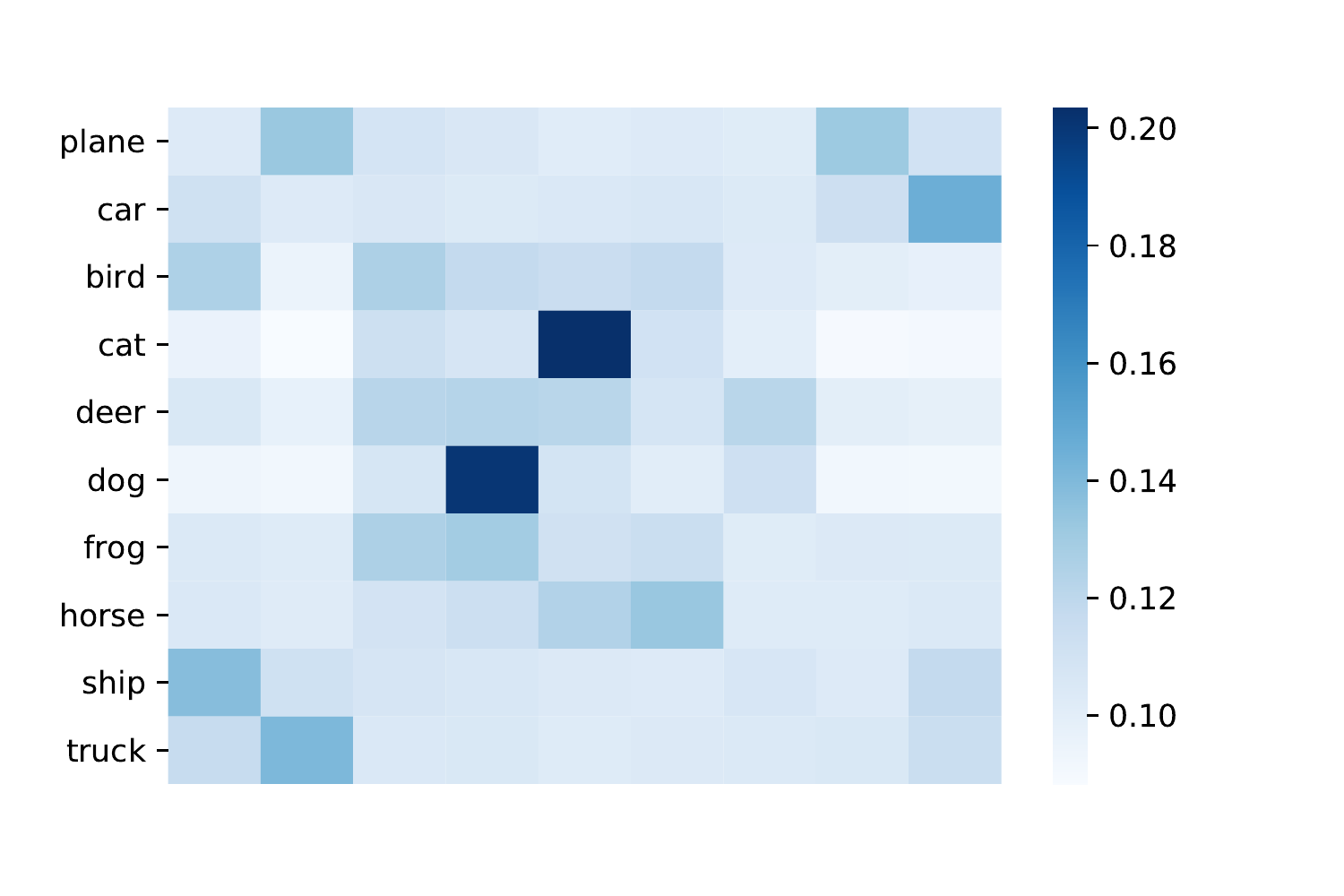}
	}
	\subfigure[200 epochs ($\mu = 99.99\%$)]{
		\includegraphics[width=1.7in]{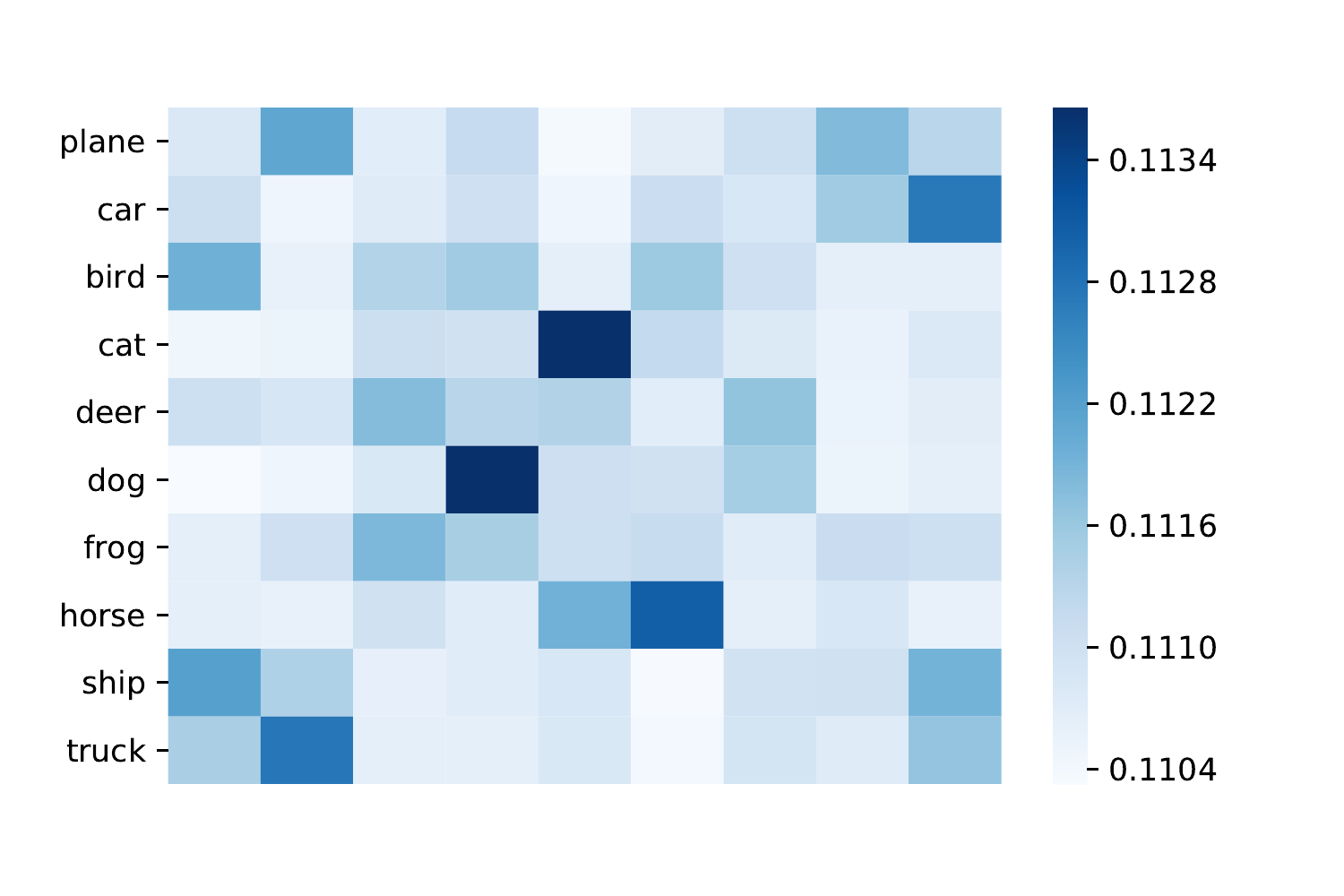}
	}
	\subfigure[300 epochs ($\mu = 99.99\%$)]{
		\includegraphics[width=1.7in]{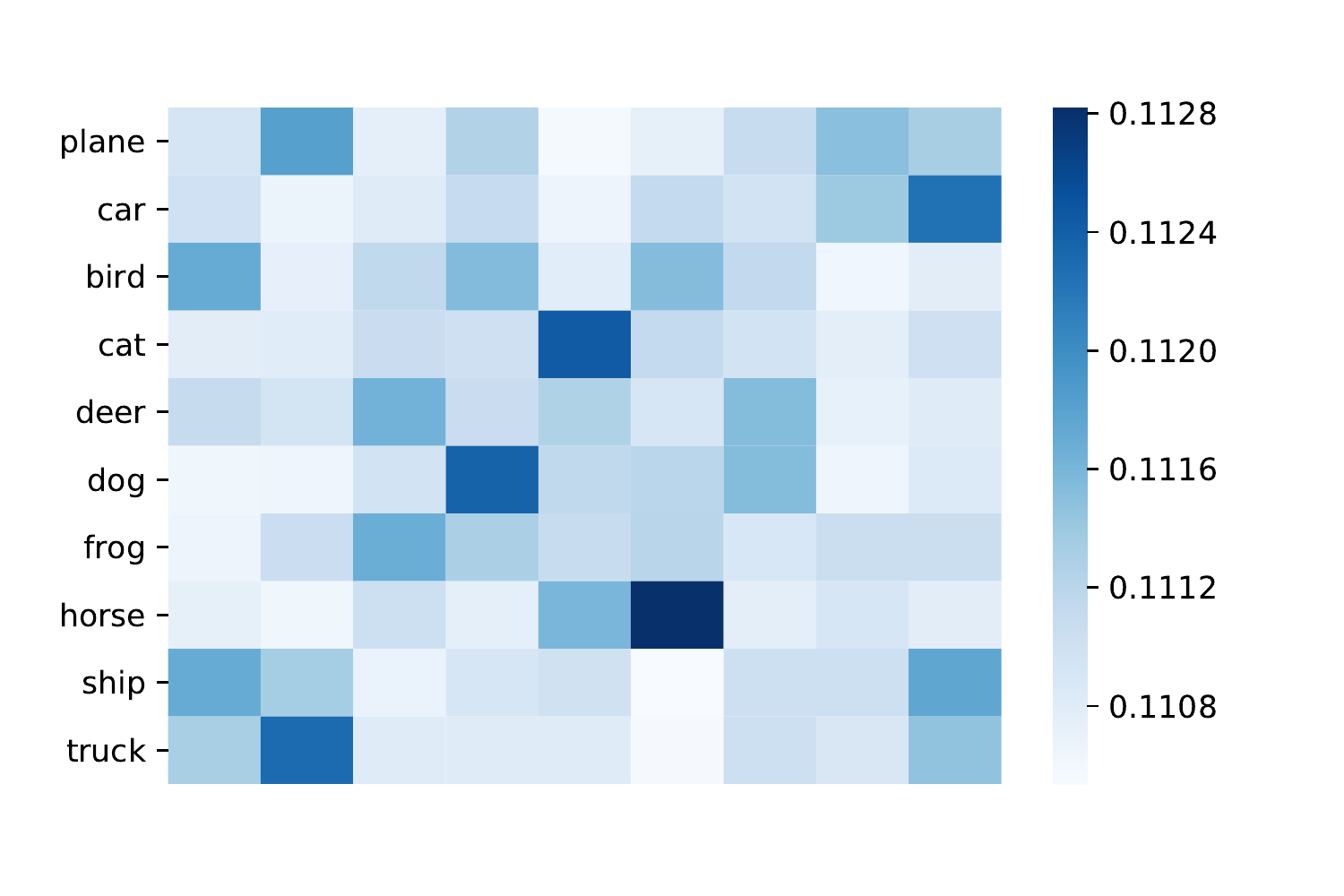}
	}
	\caption{The softmax output of residual correlation matrix on CIFAR-10: each row of them is the residual label of the corresponding class. $\mu$ represents the training accuracy after specific epochs.}
	\label{heatmap}
\end{figure}

\section{Analysis}

To explore the reason why erroneous experience can help to regularize the neural network, we visualize the softmax output of the weight of embedding layer (residual label of each class) on CIFAR-10 dataset in Figure \ref{heatmap}. According to Section \ref{reslabel}, we can know that the residual label represents the probability of ground-truth class to be wrongly classified as other classes. Our visualization results show that: (i) Residual label has consistency on time scale to some extent. For example, from the first epoch to the last epoch (the $300^{th}$ epoch), the probability of classifying the class of "cat" to "dog" (row 4, column 5) and classifying the class of "dog" to "cat" (row 6, column 4) are largest in the corresponding row of each matrix. (ii) Observing the colorbar on the right side, residual labels are becoming softer on time scale which indicates that the neural network is becoming overfitting.

We can learn from the above that: (i) Since erroneous probability is relatively stable because of the consistency of residual labels observed, the continuous training of neural network cannot make the erroneous probability change dramatically. Therefore, using the erroneous probability can regularize the neural network well. (ii) Given that the residual labels are becoming softer, in practice, the tendency of residual labels is approaching the output of the neural network according to the definition of $\mathcal{L}_{upd} $ (Eq. \ref{equ:lossupd}), which caused a lag of the output of neural networks. Simultaneously, $\mathcal{L}_{res}$  makes the neural network approach the residual label, which speeds down the overfitting. In brief, the interaction between $\mathcal{L}_{upd}$ and $\mathcal{L}_{res}$ prevent the neural network from overfitting.

\section{Conclusion}

In this paper, we propose an adaptive label regularization method for neural networks. Our method enables neural networks to learn from erroneous experience in order to prevent from overfitting. In practice, we then conduct experiments on tasks of image recognition and text classification and show the proposed method can be widely applied to all tasks with significant improvement of performance. Moreover, we show the combination of our method and the label smoothing method can work well together. And we further analyze why does our method work with pointing the consistency of residual labels. However, there is room for improvement in our approach. One is to deeply explore the reason for top-5 accuracy decreasing on ImageNet-12; another is how to extend our proposed method to more filed of deep learning except for supervised learning. In the future, given the significant improvement of our method, we hope this work will inspire more researches on label regularization.


\bibliographystyle{plain}
\bibliography{neurips_2019}

\begin{thebibliography}{10}

\bibitem{bagherinezhad2018label}
Hessam Bagherinezhad, Maxwell Horton, Mohammad Rastegari, and Ali Farhadi.
\newblock Label refinery: Improving imagenet classification through label
  progression.
\newblock {\em arXiv preprint arXiv:1805.02641}, 2018.

\bibitem{cooijmans2016recurrent}
Tim Cooijmans, Nicolas Ballas, C{\'e}sar Laurent, {\c{C}}a{\u{g}}lar
  G{\"u}l{\c{c}}ehre, and Aaron Courville.
\newblock Recurrent batch normalization.
\newblock {\em arXiv preprint arXiv:1603.09025}, 2016.

\bibitem{del2005ranking}
Gianna~M Del~Corso, Antonio Gulli, and Francesco Romani.
\newblock Ranking a stream of news.
\newblock In {\em Proceedings of the 14th international conference on World
  Wide Web}, pages 97--106. ACM, 2005.

\bibitem{goodfellow2013maxout}
Ian~J Goodfellow, David Warde-Farley, Mehdi Mirza, Aaron Courville, and Yoshua
  Bengio.
\newblock Maxout networks.
\newblock {\em arXiv preprint arXiv:1302.4389}, 2013.

\bibitem{he2016deep}
Kaiming He, Xiangyu Zhang, Shaoqing Ren, and Jian Sun.
\newblock Deep residual learning for image recognition.
\newblock In {\em Proceedings of the IEEE conference on computer vision and
  pattern recognition}, pages 770--778, 2016.

\bibitem{he2016identity}
Kaiming He, Xiangyu Zhang, Shaoqing Ren, and Jian Sun.
\newblock Identity mappings in deep residual networks.
\newblock In {\em European conference on computer vision}, pages 630--645.
  Springer, 2016.

\bibitem{hinton2015distilling}
Geoffrey Hinton, Oriol Vinyals, and Jeff Dean.
\newblock Distilling the knowledge in a neural network.
\newblock {\em arXiv preprint arXiv:1503.02531}, 2015.

\bibitem{ioffe2015batch}
Sergey Ioffe and Christian Szegedy.
\newblock Batch normalization: Accelerating deep network training by reducing
  internal covariate shift.
\newblock {\em international conference on machine learning}, pages 448--456,
  2015.

\bibitem{jaynes1957information}
E~T Jaynes.
\newblock Information theory and statistical mechanics.
\newblock {\em Physical Review}, 106(2):620--630, 1957.

\bibitem{joulin2017bag}
Armand Joulin, Edouard Grave, Piotr Bojanowski, and Tomas Mikolov.
\newblock Bag of tricks for efficient text classification.
\newblock {\em conference of the european chapter of the association for
  computational linguistics}, 2:427--431, 2017.

\bibitem{krizhevsky2009learning}
Alex Krizhevsky and Geoffrey Hinton.
\newblock Learning multiple layers of features from tiny images.
\newblock Technical report, Citeseer, 2009.

\bibitem{krizhevsky2012imagenet}
Alex Krizhevsky, Ilya Sutskever, and Geoffrey~E Hinton.
\newblock Imagenet classification with deep convolutional neural networks.
\newblock In {\em Advances in neural information processing systems}, pages
  1097--1105, 2012.

\bibitem{lai2015recurrent}
Siwei Lai, Liheng Xu, Kang Liu, and Jun Zhao.
\newblock Recurrent convolutional neural networks for text classification.
\newblock pages 2267--2273, 2015.

\bibitem{lee2014deeply}
Chen-Yu Lee, Saining Xie, Patrick Gallagher, Zhengyou Zhang, and Zhuowen Tu.
\newblock Deeply-supervised nets.
\newblock {\em arXiv preprint arXiv:1409.5185}, 2014.

\bibitem{lopez2015unifying}
David Lopez-Paz, L{\'e}on Bottou, Bernhard Sch{\"o}lkopf, and Vladimir Vapnik.
\newblock Unifying distillation and privileged information.
\newblock {\em arXiv preprint arXiv:1511.03643}, 2015.

\bibitem{micikevicius2017mixed}
Paulius Micikevicius, Sharan Narang, Jonah Alben, Gregory Diamos, Erich Elsen,
  David Garcia, Boris Ginsburg, Michael Houston, Oleksii Kuchaiev, Ganesh
  Venkatesh, et~al.
\newblock Mixed precision training.
\newblock {\em arXiv preprint arXiv:1710.03740}, 2017.

\bibitem{pereyra2017regularizing}
Gabriel Pereyra, George Tucker, Jan Chorowski, {\L}ukasz Kaiser, and Geoffrey
  Hinton.
\newblock Regularizing neural networks by penalizing confident output
  distributions.
\newblock {\em arXiv preprint arXiv:1701.06548}, 2017.

\bibitem{russakovsky2015imagenet}
Olga Russakovsky, Jia Deng, Hao Su, Jonathan Krause, Sanjeev Satheesh, Sean Ma,
  Zhiheng Huang, Andrej Karpathy, Aditya Khosla, Michael Bernstein, et~al.
\newblock Imagenet large scale visual recognition challenge.
\newblock {\em International journal of computer vision}, 115(3):211--252,
  2015.

\bibitem{srivastava2014dropout}
Nitish Srivastava, Geoffrey Hinton, Alex Krizhevsky, Ilya Sutskever, and Ruslan
  Salakhutdinov.
\newblock Dropout: a simple way to prevent neural networks from overfitting.
\newblock {\em The Journal of Machine Learning Research}, 15(1):1929--1958,
  2014.

\bibitem{srivastava2015highway}
Rupesh~Kumar Srivastava, Klaus Greff, and J{\"u}rgen Schmidhuber.
\newblock Highway networks.
\newblock {\em arXiv preprint arXiv:1505.00387}, 2015.

\bibitem{sutskever2013importance}
Ilya Sutskever, James Martens, George Dahl, and Geoffrey Hinton.
\newblock On the importance of initialization and momentum in deep learning.
\newblock In {\em International conference on machine learning}, pages
  1139--1147, 2013.

\bibitem{szegedy2016rethinking}
Christian Szegedy, Vincent Vanhoucke, Sergey Ioffe, Jon Shlens, and Zbigniew
  Wojna.
\newblock Rethinking the inception architecture for computer vision.
\newblock In {\em Proceedings of the IEEE conference on computer vision and
  pattern recognition}, pages 2818--2826, 2016.

\bibitem{vaswani2017attention}
Ashish Vaswani, Noam Shazeer, Niki Parmar, Jakob Uszkoreit, Llion Jones,
  Aidan~N Gomez, Lukasz Kaiser, and Illia Polosukhin.
\newblock Attention is all you need.
\newblock {\em neural information processing systems}, pages 5998--6008, 2017.

\bibitem{wan2013regularization}
Li~Wan, Matthew Zeiler, Sixin Zhang, Yann Le~Cun, and Rob Fergus.
\newblock Regularization of neural networks using dropconnect.
\newblock In {\em International conference on machine learning}, pages
  1058--1066, 2013.

\bibitem{wu2018group}
Yuxin Wu and Kaiming He.
\newblock Group normalization.
\newblock In {\em Proceedings of the European Conference on Computer Vision
  (ECCV)}, pages 3--19, 2018.

\bibitem{xie2016disturblabel}
Lingxi Xie, Jingdong Wang, Zhen Wei, Meng Wang, and Qi~Tian.
\newblock Disturblabel: Regularizing cnn on the loss layer.
\newblock In {\em Proceedings of the IEEE Conference on Computer Vision and
  Pattern Recognition}, pages 4753--4762, 2016.

\bibitem{yamada2018shakedrop}
Yoshihiro Yamada, Masakazu Iwamura, Takuya Akiba, and Koichi Kise.
\newblock Shakedrop regularization for deep residual learning.
\newblock {\em arXiv preprint arXiv:1802.02375}, 2018.

\bibitem{zagoruyko2016wide}
Sergey Zagoruyko and Nikos Komodakis.
\newblock Wide residual networks.
\newblock {\em arXiv preprint arXiv:1605.07146}, 2016.

\bibitem{zhang2015character-level}
Xiang Zhang, Junbo~Jake Zhao, and Yann Lecun.
\newblock Character-level convolutional networks for text classification.
\newblock {\em neural information processing systems}, pages 649--657, 2015.

\bibitem{zhang2018deep}
Ying Zhang, Tao Xiang, Timothy~M Hospedales, and Huchuan Lu.
\newblock Deep mutual learning.
\newblock In {\em Proceedings of the IEEE Conference on Computer Vision and
  Pattern Recognition}, pages 4320--4328, 2018.

\bibitem{zhu2018knowledge}
Xiatian Zhu, Shaogang Gong, et~al.
\newblock Knowledge distillation by on-the-fly native ensemble.
\newblock In {\em Advances in Neural Information Processing Systems}, pages
  7517--7527, 2018.

\bibitem{zoph2018learning}
Barret Zoph, Vijay Vasudevan, Jonathon Shlens, and Quoc~V Le.
\newblock Learning transferable architectures for scalable image recognition.
\newblock In {\em Proceedings of the IEEE conference on computer vision and
  pattern recognition}, pages 8697--8710, 2018.

\end{thebibliography}
\small
	
\end{document}